\documentclass{article}
\usepackage[margin=0.85in]{geometry}
\usepackage{srcltx}
\usepackage{amsmath}
\usepackage{amssymb}
\usepackage{amsfonts}
\usepackage{latexsym}
\usepackage{textcomp}
\usepackage{multirow}
\usepackage{booktabs}
\usepackage{url}
\usepackage{array}
\usepackage{marvosym}
\usepackage[table]{xcolor}
\usepackage{graphics}
\usepackage{graphicx}
\usepackage[caption=false]{subfig}
\usepackage{adjustbox}
\usepackage{authblk}
\usepackage{epsfig}
\usepackage[square,numbers]{natbib}
\usepackage{algorithmic}
\usepackage[lined,boxed,ruled,commentsnumbered]{algorithm2e}
\title{User-Oriented Summaries Using a PSO Based Scoring Optimization Method}

\author[1]{Augusto Villa-Monte}
\author[1]{Laura Lanzarini}
\author[2]{Aurelio F. Bariviera \thanks{aurelio.fernandez@urv.cat}}
\author[3]{Jos\'e A. Olivas}
\affil[1]{\scriptsize  Institute of Research in Computer Science LIDI (UNLP-CIC), \\ \scriptsize School of Computer Science, National~University of La Plata,\\ \scriptsize La Plata 1900, Argentina}
\affil[2]{\scriptsize  Universitat Rovira i Virgili, Department of Business, Av. Universitat 1, 43204 Reus, Spain}
\affil[3]{\scriptsize  Department of Information Technologies and Systems,\\ \scriptsize University of Castilla-La Mancha, 13071~Ciudad~Real, Spain}

\begin{document}

\maketitle

\begin{abstract}
Automatic text summarization tools have a great impact on many fields, such as medicine, law, and scientific research in general. As information overload increases, automatic summaries allow handling the growing volume of documents, usually by assigning weights to the extracted phrases based on their significance in the expected summary. Obtaining the main contents of any given document in less time than it would take to do that manually is still an issue of interest. In~this~ article, a new method is presented that allows automatically generating extractive summaries from documents by adequately weighting sentence scoring features using \textit{Particle Swarm Optimization}. The key feature of the proposed method is the identification of those features that are closest to the criterion used by the individual when summarizing. The proposed method combines a binary representation and a continuous one, using an original variation of the technique developed by the authors of this paper. Our paper shows that using user labeled information in the training set helps to find better metrics and weights.  The empirical results yield an improved accuracy compared to previous methods used in this field.
\end{abstract}

\section{Introduction}

Many years after forecasting that more information than it would be possible to process would be produced, access to information and information processing became an essential need both for academics as well as for companies and organizations. The advances in technology achieved in recent times favored the generation of large volumes of data and, as a result, the development and application of intelligent methods capable of automating their handling have become essential.

{Text documents are still the most commonly used in today's digital society \citep{schreibman2016new}. More digital textual information is being consumed every day \citep{johnson2011information}. Humans are unable to store all this information because our memory capacities are limited. We unconsciously try to retain essential information from all that information. This task, in the case of texts, is known as ``summarizing'', a cognitive characteristic of human intelligence that is used to keep what is essential. To summarize is to identify what is essential for a given purpose in a given context. Selecting relevant information is sometimes a more or less objective process, but, in many cases, it depends on the specific characteristics of the person summarizing the information. Many texts, especially those that are non-academic or non-scientific, can be examined from different points of view and therefore different essential elements. For example, the need to access and share knowledge in medicine is becoming increasingly more evident \citep{LI2006668}.}

{Basically, there are two ways to generate automatic summaries of texts: extractive, selecting the most relevant phrases, and abstractive, using an intermediate representation, such as a graph and verbalize it by generating new expressions in natural language. In the case of biomedical documents, extractive \citep{MISHRA2014457} is usually used. Obtaining such a summary can be considered as a classification problem that has two unique classes. Each phrase is labeled as ``correct'' if it is going to be part of the abstract, or ``incorrect'' if it is not \citep{Neto:2002:ATS}.  Recent works consider the task of producing extractive summaries as an optimization problem, where one or more target functions are proposed to select the ``best'' phrases from the document to form part of the summary \citep{Gambhir2017}. However, these papers consider a set of metrics that are defined a priori, and the selection of metrics is not part of the optimization process, as for example in \citep{MEENA2015244}.}

Optimization, in the sense of finding the best solution--or at least an acceptable one--for a given problem, is still a highly significant field. We are constantly solving optimization problems, for instance, when we look for the fastest way to a certain location, or when we try to get things done in as little time as possible. \textit{Particle Swarm Optimization}, which is the basis for the method proposed here, is a metaheuristic that, since its inception in 1995, has been successfully used in the resolution of a wide range of problems.

In this research work, a new method using this technique is presented, aimed at producing extractive summaries. The goal of the method is learning the ``criterion'' used by a person when summarizing a set of documents. Thus, such criterion could be applied to other documents and offer as a response a number of summaries that are similar to the ones that the person would have produced manually, but in less time. In the following sections, this ``criterion'' will be discussed in detail.

The rest of the article is organized as follows: Section \ref{sec2a} introduces an overview of the theoretical framework; Section \ref{sec2b} describes articles related to aspects discussed in this paper; Section \ref{sec3} describes the method proposed; Section \ref{sec4} includes details of the methodology used for the experiments and the results obtained; Section \ref{sec6} presents conclusions and future lines of work; and, finally, acknowledgments and references are included.

\section{Theoretical Framework} \label{sec2a}

Summarization is obtaining a number of brief, clear and precise statements that give the essential and main ideas about something. The automatic generation of text summaries is the process through which a ``reduced version'' with the relevant content of one or more documents is created using a computer \citep{torres2014automatic}.

In 1958, Luhn was first to develop a simple summarization algorithm. Since then, it has gone through constant development using different approaches, tools and algorithms \citep{Gambhir2017}. Extractive summaries are formed by ``parts'' of the document that were appropriately selected to be included in the summary. Abstractive summaries, on the other hand, are based on the ``ideas'' developed in the document and do not use the exact phrases from the original document; instead, they involve re-writing the text.

An extractive process is easier to create than an abstractive one, since the program does not have to generate new text to provide a greater level of generalization. If we consider the additional linguistic knowledge resources required to create a summary through abstraction (such as ontologies, thesaurus and dictionaries), the cost of the extractive approach is lower \citep{mani2001automatic}. An extractive summary is formed by parts of the text (from isolated words to entire paragraphs) literally copied from the source document with no complex semantic analysis \citep{Hahn:2000:CAS}. However, to successfully produce it, each part of the document must be assigned a score that represents its importance \citep{Nenkova2012}. This score allows ordering on a list, from highest to lowest, all parts of the document whose first positions are more relevant \citep{Edmundson:1961:AAI}. Finally, the summary is created using the best $n$ parts found at the top of the list.

While the extractive approach does not guarantee the narrative coherence of the sentences selected, these types of summaries reduce the size of the document, thus providing three advantages:
\begin{enumerate}
	\item[(1)] the size of the summary can be controlled,
	\item[(2)] the content of the summary is obtained accurately, and 
	\item[(3)] it can easily be found in the source document. 
\end{enumerate}

In the literature, there is abundant bibliography related to extractive summaries aimed primarily at reducing the size while keeping the information of the original document. Despite the fact that there are different extractive approaches, there is a set of metrics that is commonly used to characterize the documents and build an intermediate representation of them \citep{Lit10a}. Each metric analyzes a given characteristic of the document and allows for applying certain classification criteria to document contents. These metrics take place after the document is pre-processed, which involves the following tasks: splitting the document into sentences, tokenizing each of them, discarding stop words, applying~stemming, etc.

Table \ref{tab:metricas} shows a detail of the set of metrics that are most commonly mentioned in the literature. This table details how to calculate each of the metrics, where: $s$ is a sentence in document $d$; $D$ is the number of documents in the corpus; $S$ is the total number of sentences in document $d$; $i$ is an integer between $[0,\,S]$ sequentially assigned to each sentence from beginning to end, based on their location within the document; $|.|$ is the cardinality of the set if characters, words or key words for the text involved; and $w_i$ is the longest segment in a sentence between two key words.

\begin{table}
\caption{Set of metrics considered in this article.}
\label{tab:metricas}
\centering

\begin{tabular}{cc}
\toprule
\textbf{Type} 	& \textbf{Formula}\\
\midrule\
\multirow{3}{*}{Position}		& $i$\\
& $i^{-1}$\\
& $\max{(i^{-1},(S-i+1)^{-1})}$\\
\midrule\\
\multirow{2}{*}{Length}		& $|words(s_i)|$\\
& $|characters(s_i)|$\\
\midrule\\
\multirow{3}{*}{Keywords}		&  $\frac{|keywords(s_i)|^2}{|w_i|}$\\
& $\sum_{k\in{keywords(s_i)}}{TF(k)}$\\
& $\frac{|keywords(s_i)|}{|keywords(d)|}$\\
\midrule\\
\multirow{2}{*}{Frequency}		& $\frac{\sum_{w\in{words(S_i)}}{TF_w}}{|words(S_i)|}$\\
& $\sum_{w\in{words(S_i)}}{TF(w) \times ISF(w)}$\\
\midrule\\
\multirow{3}{*}{Title}		& $\frac{|words(s_i) \cap words(t_i)|}{\min{\big(|s_i|,|t_i|\big)}}$\\
& $\frac{|words(s_i) \cap words(t_i)|}{|words(s_i) \cup words(t_i)|}$\\
& $\frac{\vec{s_i} \times \vec{t_i}}{|\vec{s_i}| \times |\vec{t_i}|}$\\
\midrule\\
\multirow{3}{*}{Coverage}		& $\frac{|words(s_i) \cap words(d-s_i)|}{\min{\big(|s_i|,|d-s_i|\big)}}$\\[8pt]
& $\frac{|words(s_i) \cap words(d-s_i)|}{|words(d)|}$\\[8pt]
& $\frac{\vec{s_i} \times \vec{d-s_i}}{|\vec{s_i}| \times |\vec{d-s_i}|}$\\
\bottomrule
\end{tabular}
\end{table}

\textit{TF(w)} is the \textit{Term Frequency} for word $w$, and it is calculated as the number of occurrences of $w$~in~$d$ divided by the number of words in the document. In some cases, to normalize the length of the document, the number of occurrences is divided by the maximum number of occurrences. \textit{ISF(w)} is the \textit{Inverse Sentence Frequency} of $w$ and it is calculated as $1-log\big(\frac{S}{SF(w)}\big)$, \textit{SF(w)} being the number of sentences that include word $w$. \textit{ISF(w)} is an adaptation of the well-known metric \textit{TF-IDF} that is used in \textit{Information retrieval} (\textit{IR}). On the other hand, both title and coverage metrics measure in terms of words the similarity between sentence $s_i$ and another text that, in the first case, is formed by all titles in the document, and in the latter, includes the sentences that are part of the rest of the document (all~sentences in $d$ except for sentence $s_i$). As it can be seen, there are three possible calculation methods based on the similarity metrics used: \textit{Overlap}, \textit{Jaccard} and \textit{Cosine}, respectively.

Currently, all types of variations are proposed. Indeed, the calculation of frequency metrics can change taking into account only the nouns instead of all words, or position metrics can change based on whether the position of the sentence is determined within the section, the paragraph or the document. However, researchers propose new metrics combining statistical methods and discourse-based methods, including, for instance, semantic analysis. In \citep{torres2014automatic}, a more thorough list of methods with source references can be found.

\section{Related Works} \label{sec2b}  

{There are many proposals for automatically summarizing large amounts of text into informative and actionable key sentences. For example, in the field of soft-computing using machine learning and sentiment analysis, the Gist system \citep{Lovinger2019} selects the sentences that best characterize the initial documents. Other approaches use metaheuristics. \citet{10.1007/978-3-030-02840-4_7} propose two metaheuristics, Global~Best Harmony Search and LexRank Graph, hybrid algorithm, trying to optimize an objective function composed by the features of coverage and diversity.  \citet{10.1007/978-3-319-19578-0_16} propose a multi-layer approach for extractive text summarization, where the first layer consists of using two techniques of extraction: scoring of phrases and similarity for eliminating redundant phrases. The second layer optimizes the results of the previous one by a metaheuristic based on social spiders, using an objective function for maximizing the sum of similarity between sentences of the candidate summary. The last layer is for choosing the best summary from the candidate summaries generated by the optimization layer. They also use a Saving Energy Function \citep{Hamou:2015:NBM:2854500.2854502}. \citet{Masoumi} propose a biogeography-based metaheuristic optimization method (BBO) for extractive text summarization. In~ \citep{Verma2019}, authors try to generate optimal combinations of sentence scoring methods and their respective optimal weights for extracting the sentences with the help of a metaheuristic approach known as teaching–learning-based~optimization.}

{\citet{10.1007/978-3-319-26832-3_33} also try to generate an extractive generic summary with maximum relevance and minimum redundancy from multi-documents. They consider four features associated with sentences and propose a metaheuristic optimization based on solution population with multiple~objective functions that take care of both statistical and semantic aspects of the documents. \citet{10.1007/978-981-13-1501-5_36}~try to explore the strengths of metaheuristic approaches and collaborative ranking. The sentences of document are scored assigning the weight to each text feature using the metaheuristic ``Jaya'' and scores the sentences by linearly combining these feature scores with their optimal weights. They also score the sentences by simply averaging the scores of each text feature. The final ranking of sentences is calculated using collaborative ranking. In a comparative study between two bio-inspired approaches based on swarm intelligence for automatic text summaries: Social Spiders and Social~Bees~\citep{Boudia2018}, two techniques, scoring of phrases and similarity, are used for eliminating redundant phrases.}

{All of these proposals attempt to optimize the use of the usual metrics to classify sentences in order to obtain extractive summaries, trying to avoid redundancies. For this purpose, several combinations of metaheuristics are used, many of them bioinspired in the behavior of ants and swarms. In these approaches, documents are usually modeled as $n$-dimensional numeric vectors based on the calculation of $n$ metrics. These vectors are then used to generate the automatic summary through a more sophisticated algorithm \citep{Nenkova2012}. In these proposals, each vector is usually calculated for all phrases, and would allow a summary to be obtained by itself without the need to combine it with the others. However, some metrics do not allow you easily to distinguish one phrase from another, as they assign the same score to several sentences. On the other hand, the set of characteristics calculated to represent the documents is usually defined a priori and remains unchanged.}

{Human beings use} several criteria when creating a summary. Designing a program that selects significant phrases automatically requires precise instructions
. Intelligent strategies that allow mimicking human summarization are needed. This could be achieved through the combination of metrics, carefully selecting which ones to use and how to weight them. The combination of metrics allows for obtaining good results \citep{Lit10a}.

In this paper, from the representation of documents using a given set of metrics, and through a mixed discrete-continuous optimization technique based on particle swarms, the main metrics will be identified, as well as their contribution to building the expected summary. The weighted combination of the subset of metrics that better approximate the summary that would have been produced by the person constitute the desired summarization ``criterion''. In \citep{7833396}, the use of classic \textit{PSO} as a solution to this problem was proposed; the experimental results obtained showed that the strategy proposed is effective. It should be noted that the method proposed is aimed at identifying the combination of metrics that best weight the sentences. However, all metrics are considered for such weighting. In this article, we propose adding to the technique the ability of selecting the most representative metrics, while establishing their level of participation in sentence weighting. Thus, the assigned score is expected to be more accurate in relation to user preferences. In the following section, the method used to achieve this goal is discussed in detail.

\section{Proposed Method for Text Summarization}  \label{sec3}

In 1995, Kennedy and Eberhart proposed a population-based metaheuristic algorithm known as \textit{Particle Swarm Optimization} (\textit{PSO}), where each individual in the population, called particle, carries out its adaptation based on three essential factors: 
\begin{enumerate}
	\item[(i)] its knowledge of the environment (fitness value), 
	\item[(ii)] its historical knowledge or previous experiences (memory), and 
	\item[(iii)] the historical knowledge or previous experiences of the individuals in its neighborhood (social knowledge). 
\end{enumerate}

In these types of techniques, each individual in the population represents a potential solution to the problem being solved, and moves constantly within the search space trying to evolve the population to improve the solutions. Algorithm \ref{alg:PSO} describes the search and optimization process carried out by the basic \textit{PSO}~method.

\begin{algorithm}[H]
	\begin{algorithmic}[1]
		\STATE initialize necessary variables and create swarm population
		\REPEAT
		\STATE adjust inertia factor value 
		\FORALL{particles in population}
		\STATE calculate particle fitness  
		\IF{fitness is better than that of the best particle}
		\STATE update best particle and save fitness 
		\ENDIF
		\ENDFOR
		\FORALL{particles in population}
		\STATE retrieve best particle from neighborhood
		\STATE update speed and modify its position
		\ENDFOR						
		\UNTIL{reaching termination condition}
		\RETURN solution of best particle in population
	\end{algorithmic}
	\caption{Pseudocode of the basic \textit{PSO} algorithm}\label{alg:PSO}
\end{algorithm}

Since its creation, different versions of this well-known optimization technique have been developed. Originally, it was defined to work in continuous spaces, so there were spatial considerations that had to be taken into account to work in discrete spaces. For this reason, \cite{Kennedy97adiscrete} defined a new binary version of the \textit{PSO} method. One of the key problems of this new method is its difficulty to change from $0$ to $1$ and from $1$ to $0$ once it has stabilized. This drove the development of different versions of binary \textit{PSO} that sought to improve its exploratory capacity.

{Obtaining the solution to many real-life problems is a difficult task. For this reason, modifying the PSO algorithm to solve complex problems is very common. There are cases where the final solution is built from several solutions obtained by combining binary and continuous versions of PSO. However, its implementation depends on the problem type and solution structure. In this vein, such combination was already used to find classification rules to improve credit scoring \citep{Lan17}.}

Using \textit{PSO} to generate an extractive summary that combines different metrics requires combining both types of \textit{PSO} mentioned above. The subset of metrics to be used has to be selected (discrete part), and the relevance of each of these metrics has to be established (continuous part). 
 
\subsection{Optimization Algorithm} \label{3c}

To move in an $n$-dimensional space, each particle $p_i$ in the population is formed by:

\begin{enumerate}
	\item[(a)] a~binary individual $BinInd$ and its best individual $BestBinInd$, both with the format $BinInd_i~=~(binInd_{i,1}, binInd_{i,2}, \dots, binInd_{i,n})$;
	\item[(b)] a continuous individual $RealInd$ and its corresponding best individual $BestRealInd$, both with the format $RealInd_i~=~(realInd_{i,1}, realInd_{i,2}, \dots, realInd_{i,n})$;
	\item[(c)] the fitness value $fit_i$ corresponding to the individual and that of its best individual  $fitBestInd_i$; and
	\item[(d)] three speed vectors, $V1$, $V2$ and $V3$, all with format $V1_i = (v1_{i,1}, v1_{i,2}, \dots, v1_{i,n})$.
\end{enumerate}

As it can be seen, the particle has both a binary and a continuous part. Speeds $V1$ and $V2$ are combined to determine the direction in which the particle will move on the discrete space, and $V3$ is used to move the particle on the continuous space. $BinInd$ stores the discrete location of the particle, and $BestBinInd$ stores the location of the best solution found so far by it. $RealInd$ and $BestRealInd$ contain the location of the particle and that of the best solution found, the same as $BinInd$ and $BestBinInd$, but they do so in the continuous space. $fit$ is the fitness value of the individual, and $fitBestInd$ is the value corresponding to the best solution found by it. In Section \ref{secFit}, the process used to calculate the fitness of a particle from its two positions (one in each space) will be described.

Then, each time the $i$th particle moves, its current position changes as follows:

\noindent\textit{Binary part}  
\begin{eqnarray}
v1_{i,j}(t+1) = wBin \cdot v1_{i,j}(t) + \varphi1 \cdot rand1_{i,j} \cdot (2 \cdot bestBinInd_{i,j}-1) \label{eq:v1:a}\\
+ \varphi2 \cdot rand2_{i,j}\cdot (2 \cdot theBestBinInd_{i,j}-1), \nonumber
\end{eqnarray}
where $wBin$ represents the inertia factor, $Rand1$ and $Rand2$ are random values with uniform distribution in $[0,\,1]$, $\varphi1$ and $\varphi2$ are constant values that indicate the significance assigned to the respective solutions found before, $bestBinInd_{i,j}$ and $theBestBinInd_{i,j}$ correspond to the $j$th digit in binary vectors $BestBinInd$ and $TheBestBinInd$ of the $i$th particle, and $TheBestBinInd$ represents the binary position of the particle with the best fitness within the environment of particle $p_i$ (local) or the entire swarm (global). As shown in Equation (\ref{eq:v1:a}), in addition to considering the best solution found by the particle, the position of the best neighboring particle is also taken into account. Therefore, the~value $theBestBinInd_{i,j}$ corresponds to the $j$th value of vector $BinInd_k$ of particle $p_k$ with a fitness value $fit_k$ higher than its fitness ($fit_i$).

It should be noted that, as discussed in \citep{Lan11b} and unlike the Binary \textit{PSO} method described in~\citep{Kennedy97adiscrete}, the~movement of vector $V1_i$ in the directions corresponding to the best solution found by the particle and the best value in the neighborhood do not depend on the current position of the particle. Then,~each~element in speed vector $V1_i$ is calculated using Equation (\ref{eq:v1:a}) and controlled using Equation~ (\ref{eq:v1:b}):

\begin{equation}
v1_{i,j}(t)= \left \{
\begin{array}{cl}
\delta1_j & \quad if\ v1_{i,j}(t) \ge \delta1_j, \\
-\delta1_j & \quad if\  v1_{i,j}(t) \leq -\delta1_j, \\
v1_{i,j}(t) & \quad otherwise,
\end{array}
\right.
\label{eq:v1:b}
\end{equation}

\begin{equation}
\delta1_j = \frac{limit1_{j,upper}-limit1_{j,lower}}{2},
\label{eq:delta1}
\end{equation}
where $v1_{i,j} \in [-limit1_j, limit1_j]$ because of the limits that keep variable values within the set range. Then, vector $V1$ is used to update the values of speed vector $V2$, as shown in Equation (\ref{eq:v2}):

\begin{equation}
v2_{i,j}(t+1) = v2_{i,j}(t) + v1_{i,j}(t+1).
\label{eq:v2}
\end{equation}

Vector $V2_i$ is controlled in a similar way as vector $V1_i$ by changing $limit1_{j,upper}$ and $limit1_{j,lower}$ by $limit2_{j,upper}$ and $limit2_{j,lower}$, respectively. This will yield $\delta2_j$ which will be used as in Equation (\ref{eq:v1:b}) to limit the values of $V2_i$. Then, the sigmoid function is applied and the new position of the particle is calculated using Equations (\ref{eq:sigmoide:a}) and (\ref{eq:sigmoide:b}):

\begin{equation}
sig(x) = \frac{1}{1+e^{-x}},
\label{eq:sigmoide:a}
\end{equation}
\vspace{6pt} 
\begin{equation}
binInd(t+1)=\left \{
\begin{array}{cl}
1, & \quad if\ rand_{i,j}<sig(v2_{i,j}(t+1)),\\
0, & \quad if\ no, \\
\end{array}
\right.
\label{eq:sigmoide:b}
\end{equation}
where $rand_{i,j}$ is a random number with uniform distribution in $[0,\,1]$. Adding the sigmoid function in Equation (\ref{eq:sigmoide:a}) radically changes how the speed vector is used to update the position of the particle.

\noindent\textit{Continuous part}
\setlength{\arraycolsep}{0.0em}
\begin{eqnarray}
v3_{i,j}(t+1) = wReal \cdot v3_{i,j}(t) + \varphi3 \cdot rand_3{i,j} \cdot (bestBinInd_{i,j}-realInd_{i,j}) \label{eq:v3:a} \\
+ \varphi4 \cdot rand4_{i,j} \cdot (theBestRealInd_{i,j}-realInd_{i,j}) \nonumber
\end{eqnarray}
and then,
\setlength{\arraycolsep}{0.0em}
\begin{equation}
realInd_{i,j}(t+1) = realInd_{i,j}(t) + v3_{i,j}(t+1),
\label{eq:indivReal}
\end{equation}
where, once again, $wReal$ represents the inertia factor, $Rand_3$ and $Rand_4$ are random values with uniform distribution in $[0,\,1]$, and $\varphi3$ and $\varphi4$ are constant values that indicate the significance assigned to the respective solutions previously found. In this case, $TheBestRealInd$ corresponds to vector $RealInd$ from the same particle from which vector $BinInd$ was taken to adjust $V1_i$ with vector $TheBestBinInd$ in Equation (\ref{eq:v1:b}). Both $V3_i$ and $RealInd_i$ are controlled by $limit3_{j,upper}$, $limit3_{j,lower}$, $limit4_{j,upper}$ and $limit4_{j,lower}$, similar to how speed vectors $V1$ and $V2$ in the binary part were controlled.

Note that, even though the procedure followed to update vectors $V2$ and $realInd$ is the same (Equations~(\ref{eq:v2}) and (\ref{eq:indivReal})), the values of $V2$ are used as argument in the sigmoid function (Equation (\ref{eq:sigmoide:a})) to obtain a value within $[0,\,1]$ that is equivalent to the likelihood that the position of the particle takes a value of $1$. Thus, probabilities within interval $[sig(limit2_{j,lower}), sig(limit2_{j,upper})]$ can be obtained. Extreme values, when mapped by the sigmoid function, produce very similar probability values, close~to $0$ or $1$, reducing the chance of change in particle values and stabilizing it.

\subsection{Representation of Individuals and Documents} \label{sec3a}

For this article, $16$ metrics described in the literature were selected. They are based on sentence location and length on the one hand, and on word frequency and word matching on the other. Then, each sentence in each document was converted to a numeric vector whose dimension is given by the number of metrics to be used, in this case, $16$. Therefore, each document will be represented by a set of these vectors whose cardinality matches the number of sentences in it.

Using \textit{PSO} to solve a specific problem requires making two important decisions. The first decision involves the information included in the particle, and the second one is about how to calculate the particle's fitness value.

In the case of document summarization, particles compete with each other searching for the best solution. This solution consists of finding the coefficients that, when applied to each sentence metrics, have a ranking that is similar to the one established by the user. Then, following the indications detailed in the previous section, each particle is formed by five vectors and two scalar values. The~dimension of the vectors will be determined by the number of metrics to use. The binary vector $BinInd$ will determine if the metric is considered or not depending on its value---$1$ means it will be considered, $0$~means it will not. Vector $RealInd$ will include the coefficients that will weight the participation of each metric in calculating the score. The three remaining vectors are speed vectors and operate as described in the previous section.

\subsection{Fundamental Aspects of Method}

The method proposed here starts with a population of $N$ individuals randomly located within the search space based on preset boundaries. However, the binary part is not initialized the same as was the continuous one. The reason for this difference will be discussed below.

During the evolutionary process, individuals move through the discrete and the continuous spaces according to the equations detailed in Section \ref{3c}. Something that should be taken into account is how to modify speed vector when the sigmoid function (Equation (\ref{eq:sigmoide:a})) is used. In the continuous version of \textit{PSO}, the speed vector initially has higher values to facilitate the exploration of the solution space, but these are later reduced (typically, proportionally to the number of maximum iterations to perform) to allow the particle to become stable by searching in a specific area identified as promising. In this case, the speed vector represents the inertia of the particle and it is the only factor that prevents it from being strongly attracted, whether by its previous experiences, or by the best solution found by the swarm. On the other hand, when the particle's binary representation is used, even though movement is still real, the result identifying the new position of the particle is binarized by the sigmoid function instead. In this case, to be able to explore, the sigmoid function must start by evaluating values close to zero, where there is a higher likelihood of change. In the case of the sigmoid function expressed in Equation (\ref{eq:sigmoide:a}), when $x$ is $0$, it returns a result of $0.5$. This is the greatest state of uncertainty when the expected response is $0$ or $1$. Then, as it moves away from $0$, either in the positive or negative direction, its value becomes stable. Therefore, unlike the work done on the continuous part, when~working with binary \textit{PSO}, the opposing procedure must be applied, i.e., starting with a speed close to $0$ and then increasing or decreasing its value.

As already seen in Equation (\ref{eq:v1:a}), and because of the reasons explained above, an inertia factor $wBin$ is used to update speed vector $V1$, similar to the use of $wReal$ for $V3$ in (\ref{eq:v3:a}). Each factor $w$ ($wBin$~and~$wReal$) is dynamically updated based on Equation (\ref{eq:iner}):

\begin{equation}
w = wStart-(wStart-wEnd)*\frac{ite}{maxIte-1}, \label{eq:iner}
\end{equation}
where $wStart$ is the initial value of $w$ and $wEnd$ its end value, $ite$ is the current iteration, and $maxIte$ is the total number of iterations. Using a variable inertia factor facilitates population adaptation. A~higher value of $w$ at evolution start allows particles to make large movements and reach different positions in the search space. As the number of iterations progresses, the value of $w$ decreases, allowing them to perform finer tuning.

The proposed algorithm uses the concept of elitism, which preserves the best individual from each iteration. This is done by replacing the particle with lowest fitness by that with the best fitness from the previous iteration. 

As regards the end criterion for the adaptive process, the algorithm ends when the maximum number of iterations (indicated before starting the process) is reached, or when the best fitness does not change (or only slightly changes) during a given percentage of the total number of iterations.

\subsection{Fitness Function Design} \label{secFit}

Learning the criterion used by a person when summarizing a text requires having a set of documents previously summarized by that person. Typically, a person highlights those portions of the text considered to be important; in a computer, this is equivalent to assigning internal labels to the corresponding sentences. Thus, each sentence in a document is classified as one of two types--``positive,'' if it is found in the summary, or ``negative,'' if it is not.

Regardless of the problem to be solved, one of the most important aspects of an optimization technique is its fitness function. Since the summarization task presented in this work involves solving a classification problem through supervised learning, the confusion matrix will be used to measure the performance of the solution found by each particle. Among the most popular metrics used for this type of tasks, the one known as \textit{Matthews Correlation Coefficient} (\textit{MCC}) was selected.

Due to the type of problem to be solved, the sum of \textit{True Positives} and \textit{False Negatives} is equal to the sum of \textit{True Positives} and \textit{False Positives}. For this reason, by not including \textit{True Negatives} in its calculation, \textit{Recall}, \textit{Precision} and \textit{F-measure} have the same value and are not useful to differentiate the quality of the different solutions. On the other hand, \textit{MCC} does consider all cells in the confusion matrix and, therefore, it maximizes the global accuracy of the classification model. As a result, no average has to be calculated for the confusion matrices corresponding to every training document. \textit{MCC}'s values range between $[-1,\,1]$, where $1$ corresponds to the perfect model and $-1$ to the worst one. Finally, the fitness corresponding to any given individual is calculated as follows:

\begin{equation}
fitness(p_i)=\sum_{d \in corpus}{\frac{|summary(d)|}{|all \ summaries|}*MCC},
\label{eq:fit}
\end{equation}
where $|.|$ indicates the number of sentences. As it can be seen in Equation (\ref{eq:fit}), the confusion matrix used to calculate the value of \textit{MCC} must be built for each particle and each document. Building this matrix involves re-building the solution represented by the particle based on its $binInd$ and $realInd$ vectors. The first vector will allow for identifying the most representative characteristics of the criterion applied by the user, and the second one will allow weighting each of them. Even though they both represent the position of the particle in each space, the binary location is considered, since this is the one controlling the remaining fitness calculation. From the continuous vector, only the positions indicated by the binary one are used.

Since several metrics are calculated for each sentence in each document, it is expected that a linear combination of these, as expressed in Equation (\ref{eq:score}), will represent the criterion applied by the user:

\begin{equation}
score(RealInd_i, S_k) = \sum_{j=1}^{n}(realInd_{i,j}*s_{k,j}), \label{eq:score}
\end{equation}
where $\sum_{j=1}^{n}(realInd_{i,j}) = 1$, $realInd_{i,j}$ being the coefficient that individual $i$ will use to weight the value of metric $j$ in sentence $k$, indicated as $s_{k,j}$. \textit{score($RealInd_i, s_k$)} is a positive integer number proportional to the estimated significance of the sentence. Since each coefficient corresponds to the $realInd_{i,j}$ value for the individual and is within interval $[-limit4_j$, $limit4_j]$, before using it for calculations, it must be scaled to $[0,\,1]$ using such limits so that metric values are not subtracted to adjust the score. However, even though using negative coefficients for calculations is pointless, \textit{PSO} requires both positive and negative values to move particles within the search space. Adding up a metric more than once is also pointless. For this reason, once the values have been scaled, they are divided by the cumulative total to establish their individual significance in relation to the total and thus identify the metrics that have a greater influence on score calculations. As the coefficient increases, so does the significance of the metric when summarizing.

Each particle evolves to find coefficients such that, when multiplied by the values of each metric for all sentences, they allow for approximating the summary produced by the user. Once the score of all sentences in the document has been calculated, they can be sorted from highest to lowest. Those~sentences that are assigned a score of $0$ will be interpreted as irrelevant, while those that receive higher values will be more significant. User preference for a given sentence in the document is determined by the score assigned to it by the linear combination. Then, the automatic summary of the document will be obtained by considering the best $t$ sentences, $t$ being a threshold defined a priori.

It should be noted that the assessment of individual performance is not limited to all components of vector $BinInd$ whose value is $1$, but that the binary individual is used to generate combinations. All possible combinations are generated by selecting a metric for each type. The only case when no metric is used is when all positions are at $0$. When there is a single $1$ among its dimensions, the~metric corresponding to that dimension is the only one of that type that participates in combinations. This~procedure not only allows for reducing the dimensionality, but it also helps prevent inconsistencies and redundancies among metrics included in the summarization criterion. For instance, there would be no point in simultaneously using two position metrics, one that assigns a higher weight to sentences found at the end of the document and another one that does exactly the same with sentences at the beginning of the document. In this case, the method should select the position metric that assigns high values to sentences located on either end of the document. After evaluating all combinations, that with the highest fitness value is selected. As a result, vector $BinInd_i$ becomes vector $FitInd_i$ and~keeps the value indicated in $realInd_{i,j}$ only for relevant characteristics; all others are set to $0$. To~avoid excessively affecting how the optimization technique operates, each element that participates in the winning combination (each $k$ in $BinInd_i$ that was canceled when storing the final combination in $FitInd_i$) will receive a $2\%$ reduction in $v1_{i,k}$ and a $25\%$ reduction in $v2_{i,k}$. Thus, the possibility that discarded dimensions are selected in the next move of the particle is reduced, but not completely voided, which allows \textit{PSO} to explore near the solution that is currently being proposed by the particle.

Finally, after evaluating each particle's fitness value, the $FitInd_i$ will show the metrics to use and the weights included in the criterion applied by the user that effectively represent the particle whose fitness value is in $fit_i$. Even though the fitness value of the particle matches the values indicated by $FitInd$, the particle is still moving in the conventional manner using the three speed vectors.

{Algorithm \ref{alg:PSO2} shows the proposed \textit{PSO} method described previously.}

\begin{algorithm}
       \hspace*{\algorithmicindent} \textbf{Input:} popSize, maxIte
	\begin{algorithmic}[1]
		\STATE initialize $\varphi1$, $\varphi2$, $\varphi3$ and $\varphi4$
		\STATE initialize $limit1_{j,upper}$, $limit1_{j,lower}$, $limit2_{j,upper}$ and $limit2_{j,lower}$ $\forall j=1..n$
		\STATE initialize $limit3_{j,upper}$, $limit3_{j,lower}$, $limit4_{j,upper}$, $limit4_{j,lower}$ $\forall j=1..n$
		\STATE create swarm population with size maxIte
		\REPEAT
		\STATE adjust inertia factor value according to Equation (\ref{eq:iner})
		\FORALL{$i=1..popSize$}
		\STATE $fitInd_i \leftarrow $ calculate particle $p_i$ fitness according to Equation (\ref{eq:fit})		\IF{$fitInd_i > fitBestInd_i$}
		\STATE $fitBestInd_i = fitInd_i$ 
		\ENDIF
		\ENDFOR
		\FORALL{$i=1..popSize$}
		\STATE retrieve best particle from neighborhood
		\STATE update speed according to Equations (\ref{eq:v1:a})--(\ref{eq:v2}) and (\ref{eq:v3:a}) to binary and continuous part respectively
		\STATE modify particle $p_i$ position according to Equations (\ref{eq:sigmoide:b}) and (\ref{eq:indivReal})
		\ENDFOR						
		\UNTIL{reaching $maxIte$ iterations}
		\RETURN solution of best particle in population
	\end{algorithmic}
    \hspace*{\algorithmicindent} \textbf{Output:} The particle with the last best fitness value
	\caption{Proposed \textit{PSO} algorithm}\label{alg:PSO2}
\end{algorithm}

\section{Experiments and Results}  \label{sec4}

To assess the quality of the automatic summary produced by the method proposed, the summary obtained was compared with the expected one (produced by a human being) individually (on a per-document basis). To do this, freely available research articles published at a well-known medical journal were used. The documents were downloaded free of charge from the \textit{PLOS Medicine} website in \textit{XML} format.

\begin{figure}
	\centering
	\includegraphics[width=6in]{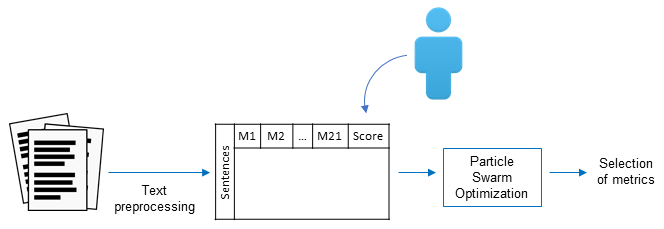}
	\caption{Methodology proposed for the summarization process.}
	\label{fig:process}
\end{figure}

Figure \ref{fig:process} shows the methodology proposed. In each document, entire sections, such as ``References'' and ``Acknowledgments,'' were discarded, as well as figures and tables. Then, titles and paragraphs were identified. Each paragraph was split into sentences using the period as delimitator, except when the period was used in numbers and abbreviations. Then, sentences were split into words, stopwords were removed, and, finally, a stemming process was applied. Once all of these pre-processing steps were completed, each of the $16$ metrics described in Section \ref{sec2a} was calculated for each sentence, escalating their values between $[0,\,1]$ per document.

As indicated in Section \ref{secFit}, summaries are created using the coefficients for the best combination of metrics selected by the particle with the highest fitness in the entire swarm, after the evolutionary process is completed. As explained in previous sections, to apply the proposed method, a set of documents that have been summarized by the user is required. This was automatically solved by using a web application whose implementation is unknown, which is equivalent to not knowing the criterion applied by the user). After analyzing several summarization applications available online, it~was decided to use the one provided by \cite{tools4noobs},
since it was the only one that met the following requirements:
\begin{enumerate}
	\item[(1)] each sentence returned corresponds to a sentence in the document,
	\item[(2)] all sentences can be ranked,
	\item[(3)] sentence ranking is established by assigning a score to each sentence, and
	\item[(4)] it has a web interface that could be integrated.
\end{enumerate}

The corpus used consisted of the $3322$ articles published between October 2004 and June 2018. Given the volume of text information, a training process was run using the documents published each month, and each result was then tested using the documents published on the following month. The percentage to be summarized was set at $10\%$. This percentage was selected based on the results obtained in \citep{7833396}. {To reduce the computational cost of calculating fitness function with such a volume of documents, they are stored and metrics previously calculated as indicated in \citep{CACIC2018}.} On the other hand, since the result depends on population initialization, $30$ separate runs were executed for each method, using a maximum of $100$ iterations. The initial population was randomly initialized with a uniform distribution in the case of the continuous part and $0$ for the binary part. The values of $limit1$, $limit2$, $limit3$ and $limit4$ were the same for all variables. In the case of $limit1$ and $limit3$, these were $[0;\,1]$ and $[0;\,0.5]$, respectively, while $limit2$ and $limit4$ had a value of $[0;\,6]$ in both cases. Therefore, the values of speed vectors $V1$ and $V3$ were limited to ranges $[-0.5,\,0.5]$ and $[-0.25,\,0.25]$, while~those of $V2$ and $realInd$ were both between $[-3,\,3]$. Population size $10$ particles in all cases. However, a~variable population strategy could be used. As regards each particle's social knowledge, global \textit{PSO} was used. The results obtained with the method proposed here are compared with those obtained with the method in \citep{7833396}, which was achieved by re-doing the tests performed then with the data used in this experiment. To do this, the parameter values used were the same as those used for the continuous part of the proposed method.

\begin{figure}
	\centering
	\includegraphics[width=6in]{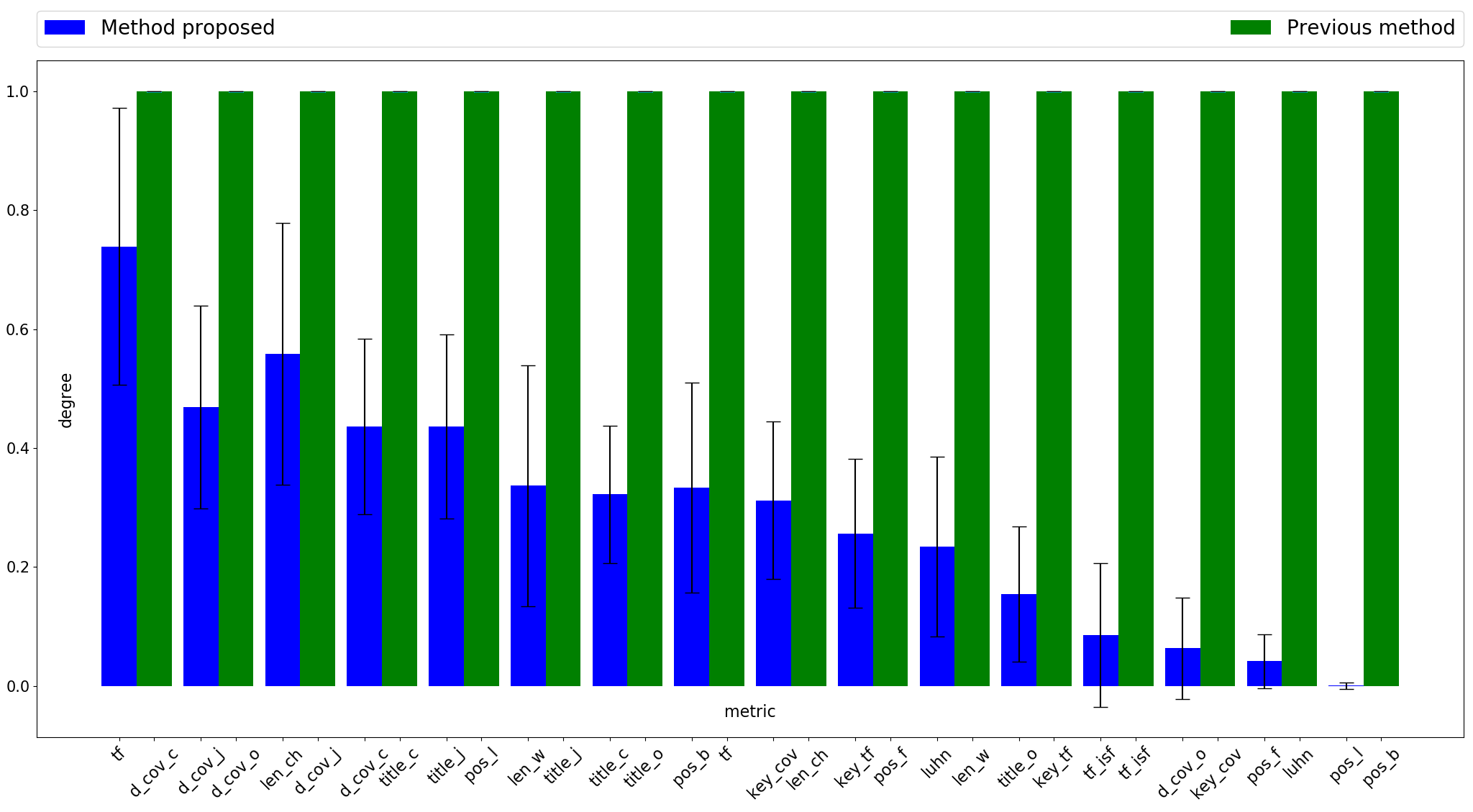}
	\caption{Participation level of metrics sorted in descending order by coefficient value.}
	\label{fig:ranking}
\end{figure}

Figure \ref{fig:ranking} shows the level of metric participation for the two methods being assessed, sorted in decreasing average coefficient order as indicated in Table \ref{tab:ranking}. These coefficients are those used to weight the value of each metric to obtain a score for each sentence. {Its value is calculated by averaging the number of times the metric is selected by the obtained particle, as in Algorithm \ref{alg:PSO2} output, among the $30$ runs performed.} For example, considering the three first values in column ``Proposed Method'' in Table \ref{tab:ranking}, it can be seen that the corresponding metrics are ``tf'', ``d\_cov\_j'' and ``len\_ch,'' whose average coefficients are $0.15$, $0.13$ and $0.13$, respectively. Therefore, the criterion indicated in Equation (\ref{eq:score}) 
 {does not distinguish between} ``d\_cov\_j'' and ``len\_ch''. However, looking at Figure \ref{fig:ranking}, it can be seen that the level of participation of ``len\_ch'' is higher than that of ``d\_cov\_j''. This is because the  {first} has been selected more times by the optimization technique. On the other hand, metric ``tf'' has the highest average coefficient for the method proposed in Table \ref{tab:ranking}, and also the highest level of participation in Figure \ref{fig:ranking}.

\begin{figure}
	\centering
	\includegraphics[width=5.2in]{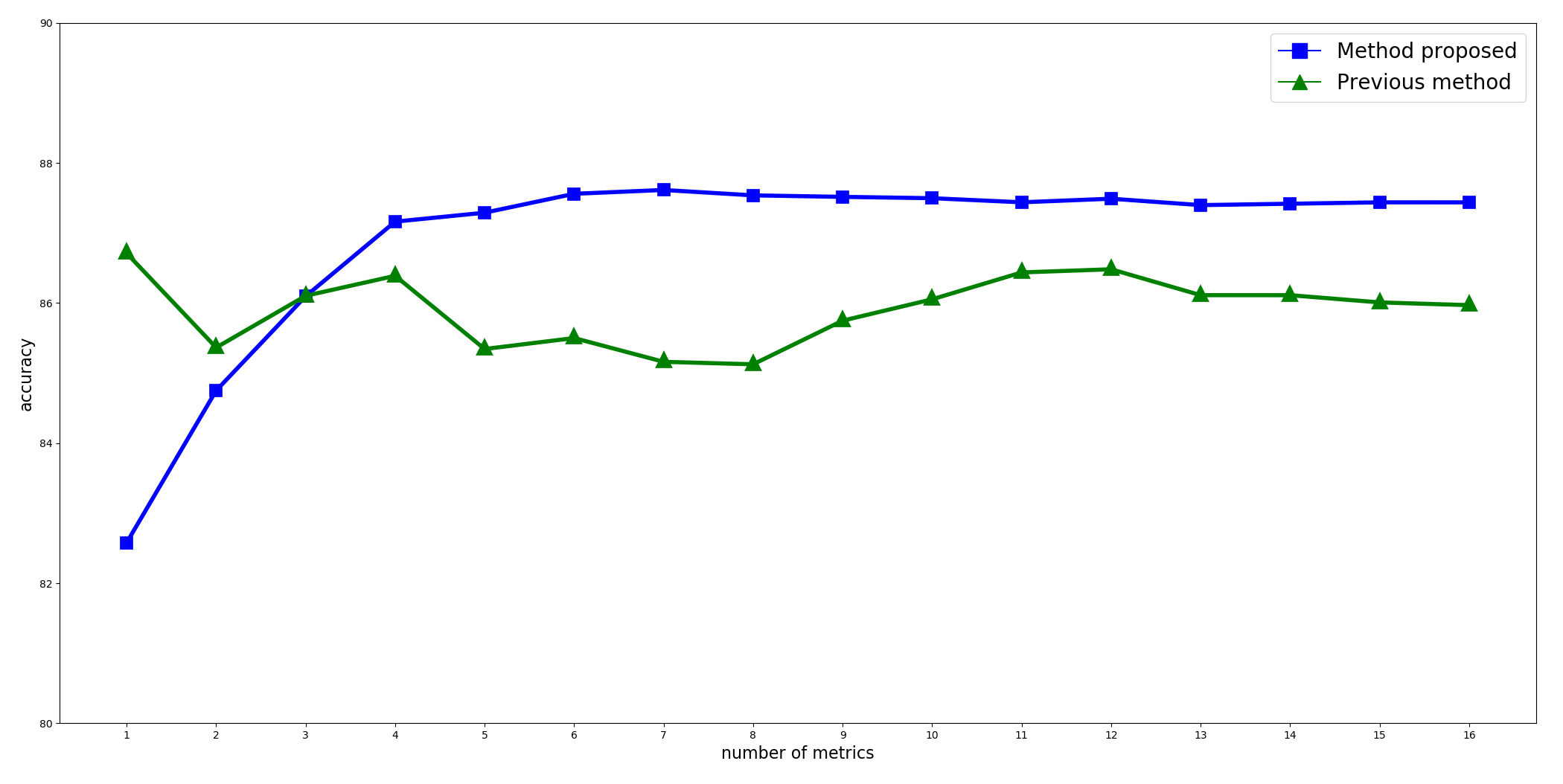}
		\caption{Accuracy evolution as new metrics are added to score calculation. {Accuracy is calculated as the ratio of selected statements by both the proposed method and the user, to the total number of corpus statements.}}
		\label{fig:accuracy}
\end{figure}

Figure \ref{fig:accuracy} shows how the accuracy of each of the methods evolves as new metrics are added. This is done in the order indicated in Table \ref{tab:ranking}. As it can be seen, the most stable behavior is that of the method proposed here. Additionally, after adding the fourth metric, accuracy becomes remarkably better than that obtained from the method in \citep{7833396}. Even though the maximum value is observed with the participation of seven metrics, four would be enough to obtain a good performance. It should also be noted that using the method described in \citep{7833396}, even if the resulting accuracy is greater for the two first metrics, the remaining ones yield a poorer result compared to the method proposed, never going above the high value of $87.61\%$.

\begin{table}
\caption {Importance of metrics in the sentence selection process, according to the respective average coefficients obtained with each method. The differences between the average values are due to the number of metrics used in each case.}
\label{tab:ranking}	
\centering
{
\begin{tabular}{cccc}
\toprule
\multicolumn{2}{c}{\textbf{Method Proposed}}&\multicolumn{2}{c}{\textbf{Previous Method}}\\
\midrule
\textbf{Metric} & \textbf{Mean $\pm$ Std. Dev.} & \textbf{Metric} & \textbf{Mean $\pm$ Std. Dev.}\\
\midrule
				tf&$0.15 \pm 0.06$&d\_cov\_c&$0.97 \pm 0.02$
				\\
				d\_cov\_j&$0.13 \pm 0.06$&d\_cov\_o&$0.97 \pm 0.03$
				\\		
				len\_ch&$0.13 \pm 0.06$ & d\_cov\_j& $0.96 \pm 0.03$ 
				\\
				d\_cov\_c&$0.12 \pm 0.05$&title\_c&$0.96 \pm 0.03$
				\\
				title\_j&$0.08 \pm 0.04$&pos\_l&$0.95 \pm 0.03$
				\\
				len\_w&$0.08 \pm 0.05$&title\_j&$0.95 \pm 0.03$
				\\
				title\_c&$0.06 \pm 0.03$&title\_o&$0.95 \pm 0.03$
				\\
				pos\_b&$0.06 \pm 0.04$&tf&$0.95 \pm 0.03$
				\\
				key\_cov&$0.05 \pm 0.03$&len\_ch&$0.95 \pm 0.03$
				\\
				key\_tf&$0.04 \pm 0.03$&pos\_f&$0.95 \pm 0.03$
				\\
				luhn&$0.04 \pm 0.03$&len\_w&$0.94 \pm 0.04$
				\\
				title\_o&$0.03 \pm 0.02$&key\_tf&$0.92 \pm 0.04$
				\\
				tf\_isf&$0.01 \pm 0.02$&tf\_isf&$0.92 \pm 0.04$
				\\
				d\_cov\_o&$0.01 \pm 0.02$&key\_cov&$0.92 \pm 0.04$
				\\
				pos\_f&$0.01 \pm 0.01$&luhn&$0.90 \pm 0.05$
				\\
				pos\_l&$0.00 \pm 0.00$&pos\_b&$0.89 \pm 0.05$
				\\
\bottomrule
\end{tabular}
}
\end{table}

Finally, the method proposed here is capable of identifying the significance of each metric at the moment of simulating user criterion. This is evident from the stability achieved in accuracy after the fourth metric is added, as observed in Figure \ref{fig:accuracy}, as well from the magnitude of the coefficients listed in Table \ref{tab:ranking}.

\section{Conclusions and Future Work} \label{sec6}

{The research carried out in this article is based on previous works such as \cite{Lit10a}, which makes evident the capacity of metrics to select sentences, even in different languages. For this reason, the~emphasis of this article is on identifying the most representative metrics to extract sentences according to human reader criteria.}

In this article, we have presented a new method for obtaining user-oriented summaries using a sentence representation based on a scoring feature subset and a mixed discrete-continuous optimization technique. It allows for automatically finding, from training documents labeled by the user, the metrics to be used and the optimal weights to summarize documents applying the same criterion.

The results obtained confirm that the selected metrics yield an adequate accuracy, being weighted as indicated by the best solution obtained using the proposed optimization technique. The tests carried out with the proposed method yielded better results than those previously established for a wide set of scientific articles from a well-known medical journal.

{One of the key features of the proposed method is its ability to reach good levels of accuracy, considering only a few metrics. In fact, the marginal contribution of additional metrics beyond five is rather low.} 

In the future, we will expand the set of metrics used to characterize input documents to obtain a richer representation, and we will also carry out tests with various summary sizes.

\end{document}